\documentclass[10pt,twocolumn,letterpaper]{article}

\usepackage{cvpr}              

\usepackage{graphicx}
\usepackage{amsmath}
\usepackage{amssymb}
\usepackage{booktabs}
\usepackage[accsupp]{axessibility}  

\usepackage[pagebackref,breaklinks,colorlinks]{hyperref}

\usepackage[capitalize]{cleveref}
\crefname{section}{Sec.}{Secs.}
\Crefname{section}{Section}{Sections}
\Crefname{table}{Table}{Tables}
\crefname{table}{Tab.}{Tabs.}

\usepackage{color}
\newcommand{\yr}{\textcolor{black}}
\newcommand{\tjs}{\textcolor{black}}
\newcommand{\rev}{\textcolor{black}}

\def\cvprPaperID{189} 
\def\confName{CVPR}
\def\confYear{2022}

\begin{document}

\title{LAKe-Net: Topology-Aware Point Cloud Completion \\ by Localizing Aligned Keypoints}

\author{Junshu Tang\textsuperscript{1},  Zhijun Gong\textsuperscript{1}, Ran Yi\textsuperscript{1}\footnotemark[1], Yuan Xie\textsuperscript{2}, Lizhuang Ma\textsuperscript{1,2}\footnotemark[1]
\\
\textsuperscript{1}Shanghai Jiao Tong University,
\textsuperscript{2}East China Normal University\\
{\tt\small \{tangjs, gongzhijun, ranyi\}@sjtu.edu.cn, yxie@cs.ecnu.edu.cn, ma-lz@cs.sjtu.edu.cn}
}

\maketitle

\renewcommand{\thefootnote}{\fnsymbol{footnote}}
\footnotetext[1]{Corresponding authors.}

\begin{abstract}
   Point cloud completion aims at completing geometric and topological shapes from a partial observation. However, some topology of the original shape is missing, existing methods directly predict the location of complete points, without predicting structured and topological information of the complete shape, which leads to inferior performance. To better tackle the missing topology part, we propose LAKe-Net, a novel topology-aware point cloud completion model by localizing aligned keypoints, with a novel \textbf{Keypoints-Skeleton-Shape} prediction manner. Specifically, our method completes missing topology using three steps: \textbf{1) Aligned Keypoint Localization.} An asymmetric keypoint locator, including an unsupervised multi-scale keypoint detector and a complete keypoint generator, is proposed for localizing aligned keypoints from complete and partial point clouds. We theoretically prove that the detector can capture aligned keypoints for objects within a sub-category. \textbf{2) Surface-skeleton Generation.} A new type of skeleton, named Surface-skeleton, is generated from keypoints based on geometric priors to fully represent the topological information captured from keypoints and better recover the local details. \textbf{3) Shape Refinement.} We design a refinement subnet where multi-scale surface-skeletons are fed into each recursive skeleton-assisted refinement module to assist the completion process. Experimental results show that our method achieves the state-of-the-art performance on point cloud completion.
\end{abstract}

\vspace{-6mm}
\section{Introduction}

The geometry and vision community has put huge effort into 
\yr{point cloud processing,} 
which is \yr{challenging} \yr{due to the unordered, unstructured characteristics, and complex semantics of the} \yr{point clouds}. 
\yr{However, in real applications, occlusions and insufficient lighting lead to partial scans of real shapes and degrade the performance of subsequent processing.}
\yr{Point cloud completion} focuses on predicting 
missing \yr{regions} 
\yr{from partial observations,} 
and shows its unique significance in many fundamental applications.

Recent works~\cite{PCN, TopNet, AtlasNet, FoldingNet, SK-PCN, GRNet, PMP, Snowflake, Pointr} for point cloud completion successfully utilized deep-learning methods and achieved more plausible and flexible results compared with \yr{traditional} geometric-based methods~\cite{pauly2005example, han2008bottom, shao2012interactive} and alignment-based methods~\cite{kalogerakis2012probabilistic, kim2013learning, martinovic2013bayesian}. 
\begin{figure}
    \centering
    \includegraphics[width=\linewidth]{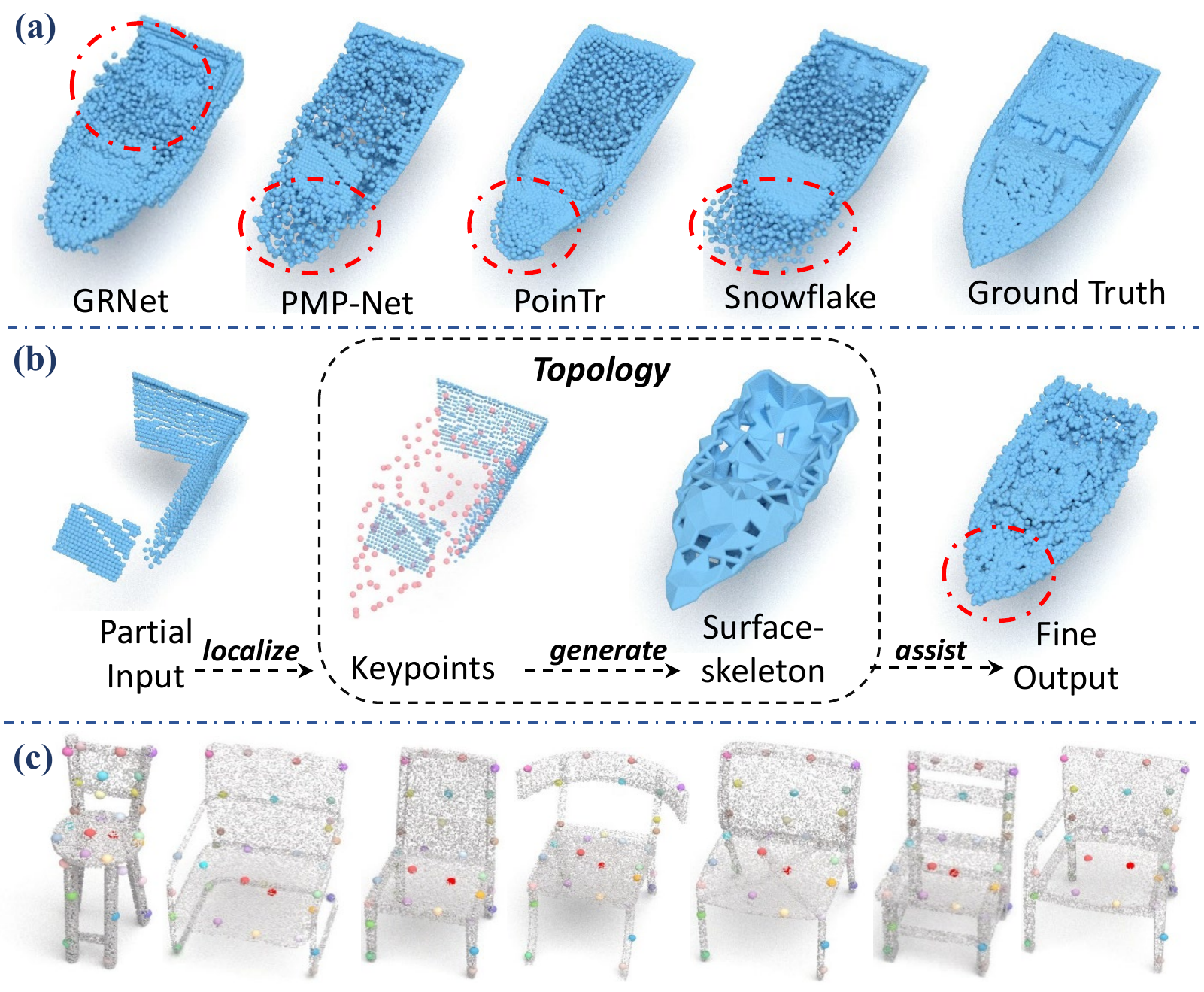}
    \vspace{-5mm}
    \caption{Illustration of (a) visual comparison results of current completion methods, (b) completion process of \yr{our} LAKe-Net, and (c) aligned keypoints. Compared with GRNet~\cite{GRNet}, PMP-Net~\cite{PMP}, PoinTr~\cite{Pointr} and Snowflake~\cite{Snowflake}, LAKe-Net can effectively recover missing topology part (see {\color{red}Red} ellipses).}
    \vspace{-5mm}
    \label{fig:fig1}
\end{figure}
However, most existing methods directly predict the location of complete points, without predicting structured and topological information of the complete shape, which leads to coarse results in missing regions (see Figure~\ref{fig:fig1}(a)).



Inspired by typical geometric modeling theory that a complete 3D object includes geometry and topology, e.g., coordinates and connectivity, we tend to predict both geometric and structured topological information for point cloud completion, including keypoints and generated skeleton. To this end, we propose a novel \textbf{Keypoints-Skeleton-Shape} prediction manner, including three steps: keypoint localization, skeleton generation, and shape refinement.

\tjs{We firstly introduce the keypoint localization.}
\yr{Different from down-sampled points,} keypoints are evenly distributed across semantic parts of the \yr{shape, and} are considered as a crucial representation of geometric structure which are widely-used in many vision applications~\cite{mian2006three, wang2018learning, you2020keypointnet}. \tjs{Therefore, we hold the belief that \textit{once the keypoints and their connectivity are correctly localized, the entire geometry is determined.} To this end, we wish to localize complete keypoints according to partial inputs under the supervision of the ground truth keypoints. However, obtaining keypoints annotation for a large number of 3D data is difficult and expensive, so we propose an \yr{asymmetric} keypoint locator including an unsupervised multi-scale keypoint detector (UMKD) and a complete keypoint generator (CKG) for complete and partial point cloud, respectively.
For UMKD,} 
we extract \textit{Aligned Keypoints}, which means the order of keypoints are the same on different objects 
within a certain category (Figure~\ref{fig:fig1}(c)), so as to represent more stable and richer information, and provide stronger supervision for predicting complete keypoints from partial inputs by CKG.

\yr{Since} discrete and sparse keypoints are not enough for representing the whole objects, 
we leverage skeleton to better represent topological details.
Inspired by existing skeleton extraction methods~\cite{cornea2007curve, cao2010point, Point2skeleton}, we 
\yr{propose a novel} {\it Surface-skeleton},
\yr{which is generated from} keypoints based on geometric priors. 
Compared with other types of skeletons, \yr{our} surface-skeleton 
is a mixture of 
curves and triangle surfaces, and can represent more complex shape information. 
\tjs{
We integrate surface-skeletons \yr{with different fineness} generated by multi-scale keypoints into the shape refinement step to recover finer results.}
\tjs{Specifically, we propose a folding-based refinement subnet including three recursive skeleton-assisted refinement modules (RSR) following some of other completion methods~\cite{SA-Net, Snowflake}.
}

In general, we propose LAKe-Net, a novel topology-aware point cloud completion model by localizing aligned keypoints.
\tjs{The whole pipeline includes four parts: auto-encoder, \yr{asymmetric} keypoints locator, surface-skeleton generator and the refinement subnet. 
\yr{We leverage pairs of complete and partial point cloud during training.}
In detail, the input point clouds (either complete or partial) are firstly fed into an auto-encoder to learn a feature embedding space and generate coarse and complete results. Then, we localize multi-scale keypoints using \yr{asymmetric} keypoints locator and generate corresponding surface-skeletons. The multi-scale structures are fed into the refinement subnet to generate fine output. The training process includes two stages for point cloud reconstruction and completion, respectively.
}

Overall, we summarize our main contributions as follows: (1) We propose LAKe-Net, a novel topology-aware point cloud completion model \yr{that utilizes a structured representation of the surface as assistance, including Aligned Keypoints and Surface-skeleton, with a new Keypoints-Skeleton-Shape prediction manner.}
(2) \tjs{We introduce an \yr{asymmetric} keypoint locator including an unsupervised multi-scale keypoint detector and a complete keypoint generator, which can capture accurate keypoints} 
\yr{for complete and partial objects in multiple categories, respectively.}
We \yr{theoretically} prove that our detector detects aligned keypoints within \yr{each} sub-category. 
(3) We conduct point cloud completion experiments on two datasets, PCN and ShapeNet55. Experimental results show that our LAKe-Net \yr{achieves the state-of-the-art performance on both datasets.} 

\begin{figure*}
    \centering
    \includegraphics[width=\linewidth]{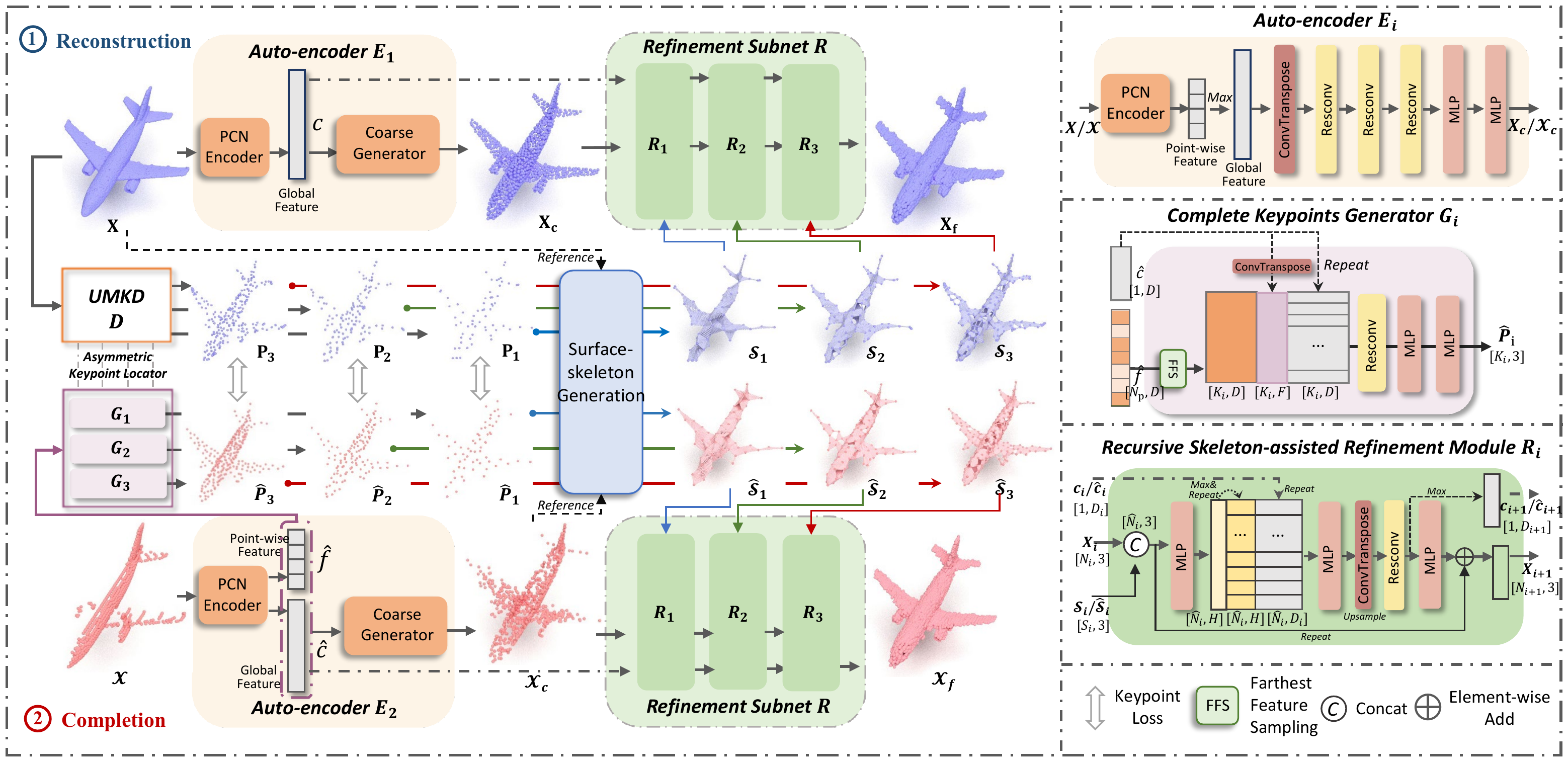}
    \vspace{-7mm}
    \caption{The overall architecture of LAKe-Net, which consists of two parts including Point Cloud Reconstruction ({\color{blue}Blue}) and Point cloud Completion ({\color{red}Red}). We show the detailed structure of (a) Auto-encoder $E$, (b) Complete Keypoint Generator $G$ and (c) Recursive Skeleton-assisted Refinement module $R$ on the right side. PCN encoder is firstly proposed in \cite{PCN}.  UMKD denotes the unsupervised multi-scale keypoint detector. Surface-skeletons $\mathcal{S}_i$ and $\hat{\mathcal{S}}_i$ are generated by $\textbf{P}_i$ and $\hat{\textbf{P}}_i$, respectively.}
    \vspace{-5mm}
    \label{fig:pipline}
\end{figure*}

\section{Related Works}

\noindent\textbf{Point Cloud Completion.}
Point cloud completion focuses on predicting missing shapes from partial point cloud input. 
Recently, inspired by point cloud analysis approaches~\cite{Pointnet, Pointnet++}, PCN~\cite{PCN} first adopts an encoder-decoder architecture and a coarse-to-fine manner to generate the complete shape. Several works~\cite{TopNet, MSN, NSFA, Snowflake} follow this practice and make modifications in network structure to obtain better performance. SA-Net~\cite{SA-Net} further extends the decoding process into multiple stages by introducing hierarchical folding. 
More recently, PoinTr~\cite{Pointr} reformulates point cloud completion as a set-to-set translation problem and designs a new transformer-based encoder-decoder for point cloud completion. 
However, these methods mostly predict the location of complete points without predicting structured and topological information, which leads to coarse results in missing regions. 
SK-PCN~\cite{SK-PCN} is the most relevant work to ours which pre-processes the dataset and uses meso-skeletons as supervision. However, SK-PCN doesn't predict the structured and topological information of original shape.
Our proposed LAKe-Net utilizes aligned keypoints and corresponding surface-skeleton which can capture the shared topological information as an assistant for completion, and obtains better performance.

\noindent\textbf{Skeleton Representation.}
Skeleton representation is widely-used in motion recognition~\cite{motion1, motion2}, human pose estimation~\cite{pose1, pose2} and human reconstruction~\cite{recon1, recon2}. Jiang \etal~\cite{jiang} propose to incorporate skeleton awareness into the deep learning-based regression for 3D human shape reconstruction from point clouds.
Tang \etal~\cite{tang2019skeleton} utilize topology preservation property of skeleton to perform 3D surface reconstruction from a single RGB image. P2P-Net~\cite{P2P-Net} learns bidirectional geometric transformations between point-based shape representations from two domains, surface-skeletons and surfaces.  
Our method designs surface-skeleton representations generated by multi-scale keypoints in a more fine-grained manner to progressively aid point cloud completion.

\noindent\textbf{Unsupervised keypoint detection.}
While most hand-crafted 3D keypoint detectors fail to detect accurate and well-aligned keypoints in complex objects, Li \etal~\cite{USIP} propose the first learning-based 3D keypoint detector USIP. However, the detected keypoints are neither ordered nor semantically salient. Fernandez \etal~\cite{fernandez2020unsupervised} utilize symmetry prior in point clouds to capture keypoints in an unsupervised manner. Recently, Jakab \etal~\cite{Keypointdeformer} further explore the application of unsupervised keypoints in shape deformation task. SkeletonMerger~\cite{SkeletonMerger} proposes a novel keypoint detector based on an autoencoder architecture. However, different models need to be trained for different categories. We propose UMKD,  an unsupervised multi-scale keypoint detector which can capture keypoints for objects in multiple categories. We find that it reaches the best performance on categories with shared topology on KeypointNet~\cite{you2020keypointnet} and produces more salient and semantic richer keypoints.

\vspace{-4mm}
\section{Proposed Method}
\vspace{-2mm}
We propose a novel topology-aware point cloud completion network by localizing aligned keypoints (LAKe-Net), whose overall architecture is shown in Fig.~\ref{fig:pipline}. The pipeline includes four parts: auto-encoder, asymmetric keypoint locator, surface-skeleton generation and shape refinement.  The training includes two stages:
point cloud reconstruction and completion.
The training data consists of pairs of complete and partial point clouds ${(\textbf{X},\mathcal{X})}$, 
where $\textbf{X}\in\mathbb{R}^{N_c\times 3}$ and $\mathcal{X}\in\mathbb{R}^{N_p\times 3}$ denote the coordinates of complete and partial point clouds, $N_c$ and $N_p$ denote the number of points of complete and partial data, respectively. 

\begin{figure}[]
    \centering
    \includegraphics[width=\linewidth]{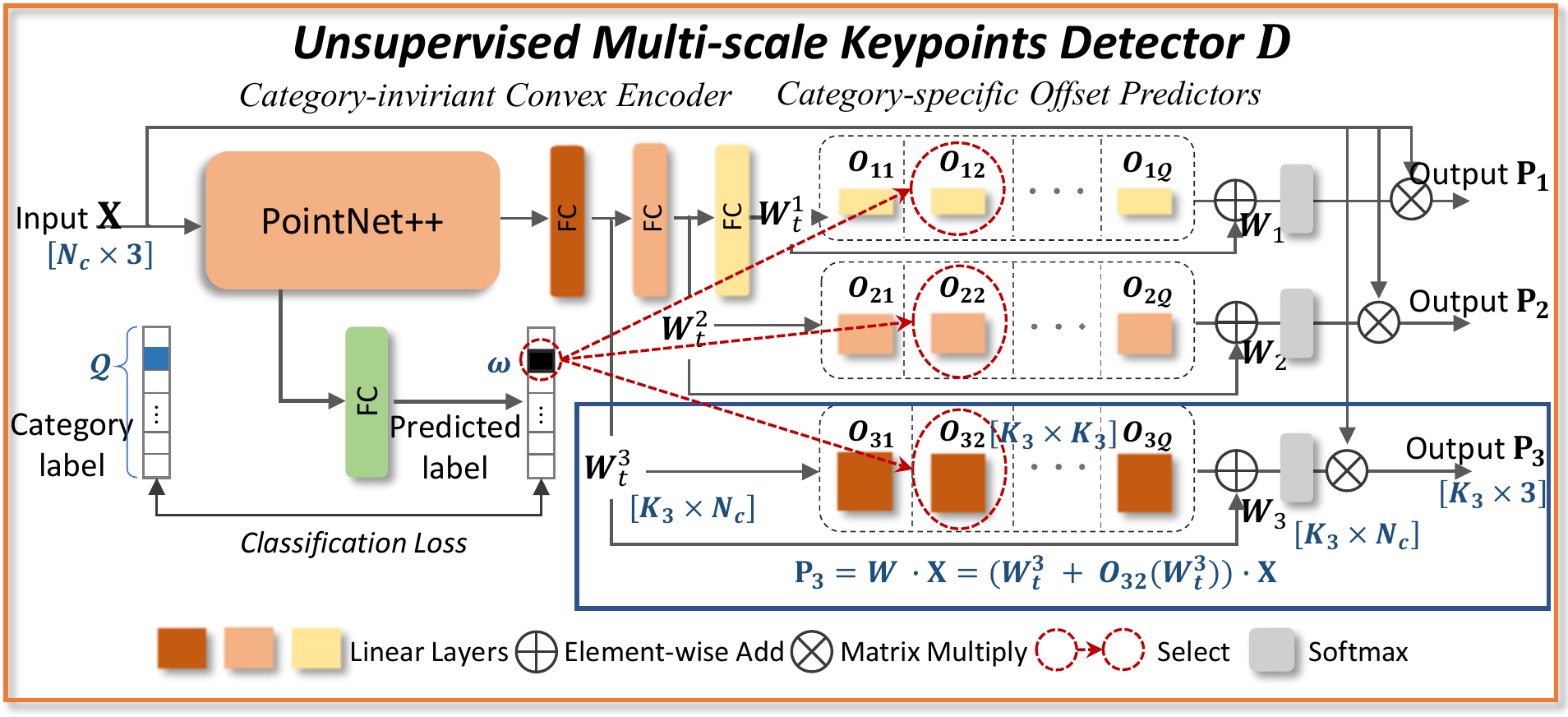}
    \vspace{-5mm}
    \caption{The detailed structure of our proposed UMKD.  Take $\textbf{P}_3$ as an example, we show the calculation process and dimensions of relative tensors in {\color{blue}Blue}.} 
    \vspace{-7mm}
    \label{fig:predictor}
\end{figure}


Firstly, see upper part of Fig.~\ref{fig:pipline}, we utilize complete data $\textbf{X}$ as input to train an unsupervised multi-scale keypoint detector (UMKD) $D$, which extracts the multi-scale keypoints $\{\textbf{P}_i\}_{i=1}^3$, and an auto-encoder $E_1$, which maps the inputs into a global feature space $c$ and obtain coarse results $\textbf{X}_c$. Then, a surface-skeleton generation process is employed to 
leverage the topology information in keypoints to construct a finer representation.
Finally, a refinement subnet adopts the topology information $\mathcal{S}$ and coarse results $\textbf{X}_c$ to generate high-resolution results $\textbf{X}_f$. \rev{The reconstruction part unsupervisedly trains the UMKD to learn aligned keypoints for later completion stage and also provides a good initialization of the network to predict complete shape.}

In the second stage, we fix the weights of keypoint detector $D$ and auto-encoder $E_1$, then input partial data $\mathcal{X}$ into a new auto-encoder $E_2$ and predict complete coarse results $\mathcal{X}_c$. It is noteworthy that the auto-encoder $E_1$ and $E_2$ have the same architecture which includes a PCN encoder~\cite{PCN} and coarse point generator. The encoder of $E_2$ embeds input points into point-wise local features $\hat{f}\in\mathbb{R}^{N_p\times d}$, where $d$ denotes the dimension of feature embedding. We consider the maximum value $\hat{c}=max_{N_p}(\hat{f})\in\mathbb{R}^{1\times d}$ as global features. Then we fuse the local and global features and feed the fused feature into a complete keypoint generator (CKG) $G$ to predict multi-scale keypoints $\{\hat{\textbf{P}}_i\}_{i=1}^3$, supervised by keypoints $\{\textbf{P}_i\}_{i=1}^3$ detected in the first stage. At last, we send $\mathcal{X}_c$ and interpolated surface-skeleton $\hat{\mathcal{S}}$ to a skeleton-assisted refinement subnet $R$ to generate fine outputs $\mathcal{X}_f$.
We describe the technical details of our proposed modules and training losses in the following sections.

\subsection{Unsupervised Multi-scale Keypoint Detector}
\label{sec:kpd}
Given a set of $N_c$ input complete point clouds $\textbf{X}=\{x_j|j=1,\cdots,N_c\}\in\mathbb{R}^{N_c\times3}$, our aim is to predict $\{K_i\}_{i=1}^3$ numbers of keypoints $\{\textbf{P}_i\}_{i=1}^{3}=\{p_k|k=1,\cdots,K_i\}\in\mathbb{R}^{K_i\times3}$. 
Specifically, we tend to predict convex combination weights $\textbf{W}_i=\{w_{ij}\}\in\mathbb{R}^{N_{c}\times K_i}$ of point clouds instead of predicting the coordinates of keypoints directly to avoid deviating from the original shape.  So the predicted keypoints $\textbf{P}_i$ are derived by: 
\begin{equation}\small
    \begin{aligned}
       \textbf{P}_i=\textbf{W}_{i}^{T}\textbf{X}= \sum^{N_c}_{j=1}w_{ij}^{T}x_{j}, s.t., w_{ij}>0, \sum^{N_c}_{j=1}w_{ij}=1.
    \end{aligned}
\end{equation}

To predict convex weight $\textbf{W}_{i}$ using a single model for all categories which adapts to our pipeline, we assume $\textbf{W}_{i}=\textbf{W}^{i}_{t} + \textbf{W}^{i}_{o}$, where $\textbf{W}^{i}_{t}$ denotes a category-invariant template weight and $\textbf{W}^{i}_{o}$ denotes a category-specific weight offset. Besides, we expect to predict multi-scale keypoints for subsequent tasks.
To this end, we propose an Unsupervised Multi-scale Keypoint Detector (UMKD) which consists of a category-invariant convex encoder and category-specific offset predictors. The detailed structure of UMKD is shown in Figure~\ref{fig:predictor}. In detail, we apply a PointNet++~\cite{Pointnet++} as a backbone encoder to extract local and global features. It includes four set abstraction layers to group and downsample input points. Then the global and local features are propagated back to each partial point. Then we input point-wise features into three fully connected blocks and extract multi-scale convex features $\textbf{W}^i_t\in\mathbb{R}^{K_i\times N_c}$ progressively. 


As for predicting category-specific offset $\textbf{W}^{i}_{o}$, we firstly predict the category label $\omega$ for input shape and send the $\textbf{W}^{i}_t$ to the relative offset predictor $\textbf{O}_{i\omega}$. We add two fully-connected layers after the last pointset abstraction layer in PointNet++ as classification head to predict the category label of every input geometries. $\textbf{O}_{ij}\in\mathbb{R}^{K_i\times K_i}$ is a learnable matrix. We set $j\in[1,\mathcal{Q}]$. $\mathcal{Q}$ denotes the number of categories. 
Take $\textbf{P}_{3}$ as an example (shown in Figure~\ref{fig:predictor}), the input points $\textbf{X}\in\mathbb{R}^{N_c\times 3}$ are send to the convex encoder and output template convex weights $\textbf{W}^{3}_{t}\in\mathbb{R}^{K_3\times N_c}$ and predicted label $\omega=2$. Then $\textbf{W}_{t}^{3}$ are input to the selected offset predictor $\textbf{O}_{32}$ and output $\textbf{W}^{3}_{o}\in\mathbb{R}^{K_3\times N_c}$. At last, we get the final convex weight $\textbf{W}_3 = \textbf{W}^{3}_{t} + \textbf{W}^{3}_{o}$ and normalize $\textbf{W}_3$ by a softmax function and predict keypoint $\textbf{P}_3$.



The keypoints extracted by UMKD $D$ follow a theory that:
\textbf{\textit{The coordinates of detected keypoints $\textbf{P}$ are irrelevant to the order of original points $\textbf{X}$. That is: P = D($\textbf{X}$) = D($\mathcal{R}(\textbf{X})$), where $\mathcal{R}(\cdot)$ denotes random permutation operation.}} The theoretical proof is introduced in Supplementary Materials. Therefore, the detected keypoints are aligned among objects with shared topology within a sub-category (shown in  Figure~\ref{fig:fig1}(c)).



\begin{figure}
    \centering
    \includegraphics[width=\linewidth]{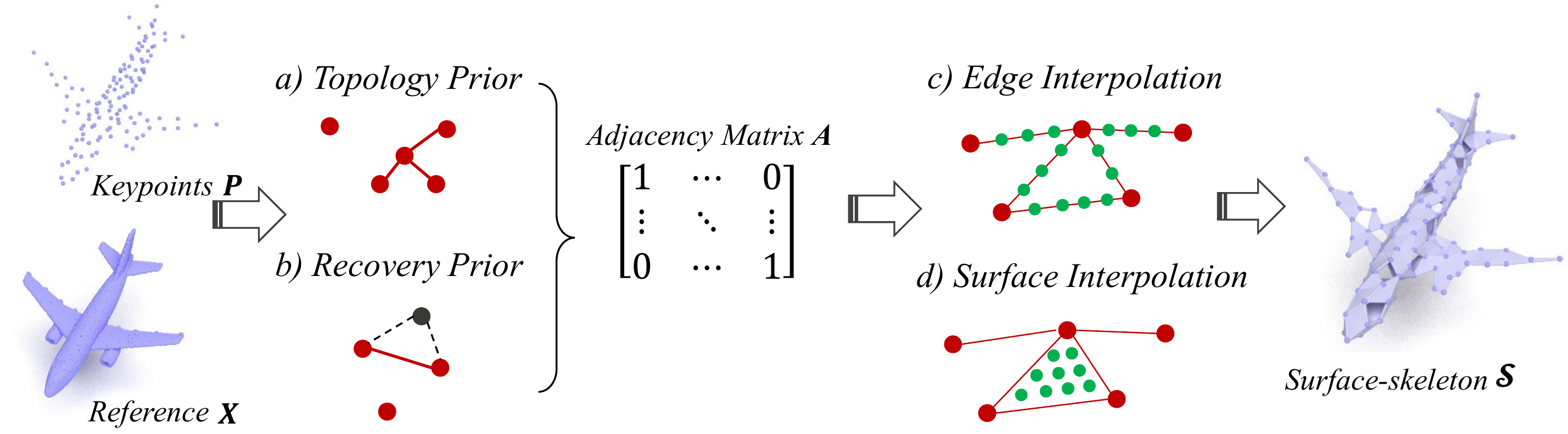}
    \vspace{-7mm}
    \caption{The surface-skeleton generation operation.}
    \vspace{-5mm}
    \label{fig:graph}
\end{figure}

\subsection{Surface-skeleton Generation}
\label{sec:mgg}
After extracting the keypoints of original point clouds, our aim is to reconstruct the point cloud according to the extracted keypoints. We consider skeleton as an intermediate representation between keypoints and original point clouds. Previous methods like SkeletonMerger~\cite{SkeletonMerger} used skeletons connected by every pair of keypoints, which leads to high computational complexity and a lot of invalid points.
\yr{Moreover, given a surface, previous skeletons either are located near the medial axis of the surface~\cite{cornea2007curve}, or are mostly located outside the surface~\cite{SkeletonMerger}.
We aim at extracting a skeleton that are located near the surface, so that it can better assist the later refinement process.}
In order to represent both topological and geometric information of a complex shape, we design a surface-skeleton structure which is generated by keypoints and consists of a mixture of curves and triangle surfaces adapted to the local 3D geometry. 
Specifically, given predicted keypoints $P\in\mathbb{R}^{K\times3}$ and reference points $X$, we follow the two shape priors proposed in~\cite{Point2skeleton} and generate the skeletal graph: (1) the topology prior that each node has links to its top-$2$ nearest nodes; (2) the recovery prior that two keypoints are linked if they are two nearest keypoints of a reference point.

After skeletal graph generation, we can get an adjacency matrix $A\in\mathbb{R}^{K\times K}$.
We propose a surface interpolation strategy based on Delaunay-based triangulation region. The whole interpolation process consists of two steps: edge interpolation and triangle surface interpolation. Given a certain graph, we firstly interpolate points into each connected edge. Next, we detect every triangle surface according to the skeletal graph. Then we insert points into the triangle region. The number of interpolated points is proportional to the triangle area. The overall skeletal graph generation and surface interpolation process is shown in Fig~\ref{fig:graph}. It is noteworthy that we utilize the complete point clouds $\textbf{X}$ as reference shape for reconstruction while using the coarse results $\mathcal{X}_c$ for completion. 

Overall, we get the surface-skeleton $\mathcal{S}$. 
Increasing the number of keypoints leads to more complex surface-skeletons which can represent finer shape details.
We utilize the structure as topology representation of geometric shape which is crucial for reconstruction and completion.

\subsection{Complete Keypoint Generator}
\label{sec:ckg}
In the second stage, given the partial input $\mathcal{X}\in\mathbb{R}^{N_p\times3}$, the aim of our proposed Complete Keypoint Generator (CKG) module is to predict multi-scale complete keypoints from partial feature embedding. To this end, we utilize the local and global feature $\hat{f}$ and $\hat{c}$ extracted by the encoder in $E_2$ as input. 
Similar to Farthest Point Sampling (FPS) strategy for point cloud, in order to downsample the point features, we use Farthest Feature Sampling (FFS) strategy to down sample point-wise local feature $\hat{f}$ to sparse feature $\hat{f}^*$ where we 
replace point coordinates in FPS by feature embeddings.
The number of sampling is the same as predicted keypoints. Then we utilize a de-convolution layer to upsample the global feature $\hat{c}$ and fuse them into a residual block and predict final keypoints. We train three similar blocks for multi-scale keypoints prediction. Then we generate corresponding surface-skeleton $\hat{\mathcal{S}}$ by surface interpolation introduced on Sec.~\ref{sec:mgg}.

\subsection{Recursive Skeleton-assisted Refinement}
\label{sec:rsr}
Our proposed shape refinement subnet $R$ includes three recursive skeleton-assisted refinement (RSR) modules that aim to integrate multi-scale surface-skeletons and coarse output from previous auto-encoder to predict finer geometric details in a recursive way.  The detailed design of the module is shown in Fig~\ref{fig:pipline}. It follows existing methods~\cite{FoldingNet, SA-Net, Snowflake} using a coarse-to-fine strategy to learn the offset of integrated points. 
Specifically, we progressively concatenate coarse point clouds obtained from previous steps and surface-skeleton generated by corresponding keypoints described in Sec.~\ref{sec:mgg}.  We denote the input coarse points as $\textbf{X}_{i-1}=\{x_j\}_{j=1}^{N_{i-1}}$ and surface-skeleton as $\mathcal{S}_{i-1}=\{p_j\}_{j=1}^{S_{i-1}}$. The integrated points $\hat{\textbf{X}}_{i-1}=concat(\textbf{X}_{i-1}, \mathcal{S}_{i-1})$ on the $i$-th step where $concat(\cdot)$ refers to concatenate operation. $N_{i-1}$ and $S_{i-1}$ denote numbers of coarse points and surface-skeleton, respectively. In this paper, we set $N_i = S_i$ in each step. Therefore, the updated points output by the $i$-th RSR module $\textbf{X}_{i} = \hat{\textbf{X}}_{i-1} + R(\hat{\textbf{X}}_{i-1})$ will be sent to the next step. 

\subsection{Training and Losses}
\noindent\textbf{Point Cloud Reconstruction.}
In the first stage, the UMKD $D$, auto-encoder $E_1$ and refinement subnet $R$ are trained together.
The training losses are divided into two parts, one is to constrain the keypoint detection, the other is data reconstruction.
Firstly, in order to encourage the detected keypoints $\textbf{P}$ to be well-distributed and not deviate from the global shape, we calculate the Chamfer Distance (CD) loss between the predicted keypoints and sparse point clouds $\textbf{X}^{*}$ downsampled from input data using FPS strategy. 
As for training one detector within several categories, we also train a classification head. We denote the predicted output is $\omega$ and certain category label is $\sigma$. 
We train a criterion loss $\mathcal{L}_{cls}$.
Besides, as mentioned in Sec.~\ref{sec:mgg}, we expect the surface-skeleton can reconstruct the geometric shape of the ground truth. We calculate CD between multi-scale surface-skeletons $\{\mathcal{S}_i\}^{3}_{i=1}$ and the ground truth $\textbf{X}$. 
So the overall loss for training keypoint detector is:
\begin{equation}\small
    \begin{aligned}
    \mathcal{L}_{CD} = \frac{1}{|X|}\sum_{x\in X}\min_{y\in Y}||x-y||^{2} + \frac{1}{|Y|}\sum_{y\in Y}\min_{x\in X}||y-x||^{2},
    \end{aligned}
    \label{eq:cd}
    \vspace{-3mm}
\end{equation}
\begin{equation}\small
    \begin{aligned}
    \mathcal{L}_{cls} = -\sum^{\mathcal{Q}}_{i=1}(\sigma_{i}log\omega_i + (1-\sigma_{i})log(1-\omega_{i})),
    \end{aligned}
\end{equation}
\begin{equation}\small
    \begin{aligned}
    \mathcal{L}_{kp} = \mathcal{L}_{CD}(\textbf{P}, \textbf{X}^{*}) + \sum^{3}_{i=1}\mathcal{L}_{CD}(\mathcal{S}_i, \textbf{X})+ \mathcal{L}_{cls}.
    \end{aligned}
\end{equation}

At last, we calculates CD between the ground truth $\textbf{X}$ and sparse output 
$\textbf{X}_c$, dense output $\textbf{X}_f$, respectively.
\begin{equation}\small
    \begin{aligned}
    \mathcal{L}_{rec} = \mathcal{L}_{CD}(\textbf{X}_c, \textbf{X}) + \mathcal{L}_{CD}(\textbf{X}_f, \textbf{X}).
    \end{aligned}
\end{equation}

In general, the overall training loss in the first stage is:
\begin{equation}\small
    \begin{aligned}
    \mathcal{L}_{1} = \mathcal{L}_{rec} + \lambda_{kp}^{1}\mathcal{L}_{kp},
    \end{aligned}
\end{equation}
where $\lambda_{kp}^{1}$ denotes hyper-parameters to balance inference.

\noindent\textbf{Point Cloud Completion.} In the second stage, we fix the weights of UMKD $D$ and auto-encoder $E_1$, and train a new auto-encoder $E_2$ and CKG $G$. The refinement subnet $R$ pre-trained before continues to be optimized. We constrain keypoints prediction using absolute distance between predicted keypoints $\hat{\textbf{P}}$ and ground truth keypoints $\textbf{P}$:
\begin{equation}\small
    \begin{aligned}
        \mathcal{L}^{c}_{kp}=\sum^{3}_{i=1}\sum^{K_i}_{j=1}||p_{ij} - \hat{p}_{ij}||^{2}.
    \end{aligned}
\end{equation}

Same as other concurrent network~\cite{ASFM}, we align global features $c$ and $\hat{c}\in\mathbb{R}^{1\times d}$ encoded by auto-encoders in two stages for hidden feature space learning:
\begin{equation}\small
    \begin{aligned}
        \mathcal{L}_{feat} =\frac{1}{d} \sum^{d}_{i=1}||c_{i}-\hat{c}_{i}||^{2}.
    \end{aligned}
\end{equation}

As for typical training on completion task, we follow the coarse-to-fine process in the first stage. The coarse output $\mathcal{X}_{c}$ and fine output $\mathcal{X}_{f}$ are optimized using CD loss:
\begin{equation}\small
    \begin{aligned}
    \mathcal{L}_{com} = \mathcal{L}_{CD}(\mathcal{X}_c, \textbf{X}) + \mathcal{L}_{CD}(\mathcal{X}_f, \textbf{X}).
    \end{aligned}
\end{equation}

In the summary, the full objective of point cloud completion in the second stage is:
\begin{equation}\small
    \begin{aligned}
    \mathcal{L}_{2} = \mathcal{L}_{com} + \lambda_{kp}^{2}\mathcal{L}_{kp}^{c} + \lambda_{feat}\mathcal{L}_{feat},
    \end{aligned}
\end{equation}
where ($\lambda^{2}_{kp}$, $\lambda_{feat}$) denote hyper-parameters.


\section{Experiments}


\subsection{Dataset Setting and Evaluation metric}
\noindent\textbf{PCN:} The PCN dataset is a widely-used benchmark for point cloud completion, which is created by~\cite{PCN}, including different objects from 8 categories: plane, cabinet, car, chair, lamp, sofa, table, and vessel. The training set contains 28,974 objects, while validation and test set contains 800 and 1,200 objects, respectively.
The complete point cloud consists of 16,384 points which are uniformly sampled on the original CAD model. Partial point cloud, consisting of 2,048 points, is created by back-projecting 2.5D depth images into 3D from 8 random viewpoints.

\noindent\textbf{ShapeNet55:} To explore the performance of our method on a large number of categories, we evaluate our method on all 55 categories of ShapeNet~\cite{Shapenet}, named ShapeNet55. The ShapeNet55 dataset was first created by PoinTr~\cite{Pointr}. The training set contains 41,952 objects, while test set contains 10,518 objects. We randomly sample 80$\%$ objects in each category to form training set and use the rest 20$\%$ to form validation set.

\noindent\textbf{Evaluation Metrics:}
We utilize two evaluation metrics between output point cloud and the ground truth, Chamfer Distance (CD) using L2 norm and Earth Mover's Distance (EMD), following most of the methods on PCN and ShapeNet55 test set. CD is introduced in Equation~\ref{eq:cd} and EMD is defined as:
\begin{equation}\small
    \begin{aligned}
    EMD(X, Y) = \min_{\phi: X\to Y}\frac{1}{|X|}\sum_{x\in X} ||x-\phi(x)||_{2},
    \end{aligned}
\end{equation}
where $\phi$ is a bijection. It is noteworthy that we compute these metrics using 16,384 and 8,192 points for PCN and ShapeNet55, respectively.

\subsection{Implementation Details}
The whole training of LAKe-Net is a two-stage process: point cloud reconstruction and point cloud completion. The input of the first stage (reconstruction) is a set of complete point clouds with coordinates and object category labels from training set of all datasets. 
We train the keypoint detection for 60 epochs and progressively extract 256, 128, 64 keypoints. The refinement subnet includes three RSR modules, the up factors of de-convolution are [1,1,2].
For the second stage (the bottom completion branch of Fig.~\ref{fig:pipline}), we only input partial point clouds from training set with its coordinate information. We utilize Adam optimization to train the whole architecture of point cloud completion for 100 epochs with batchsize 64 and learning rate 0.001. The hyper-parmeters $\lambda_{kp}^{1}=\lambda_{kp}^{2}=10$, $\lambda_{feat}=1000$. \rev{The inference time  of our method is 34.5ms per sample.}

\begin{figure}[]
    \centering
    \includegraphics[width=\linewidth]{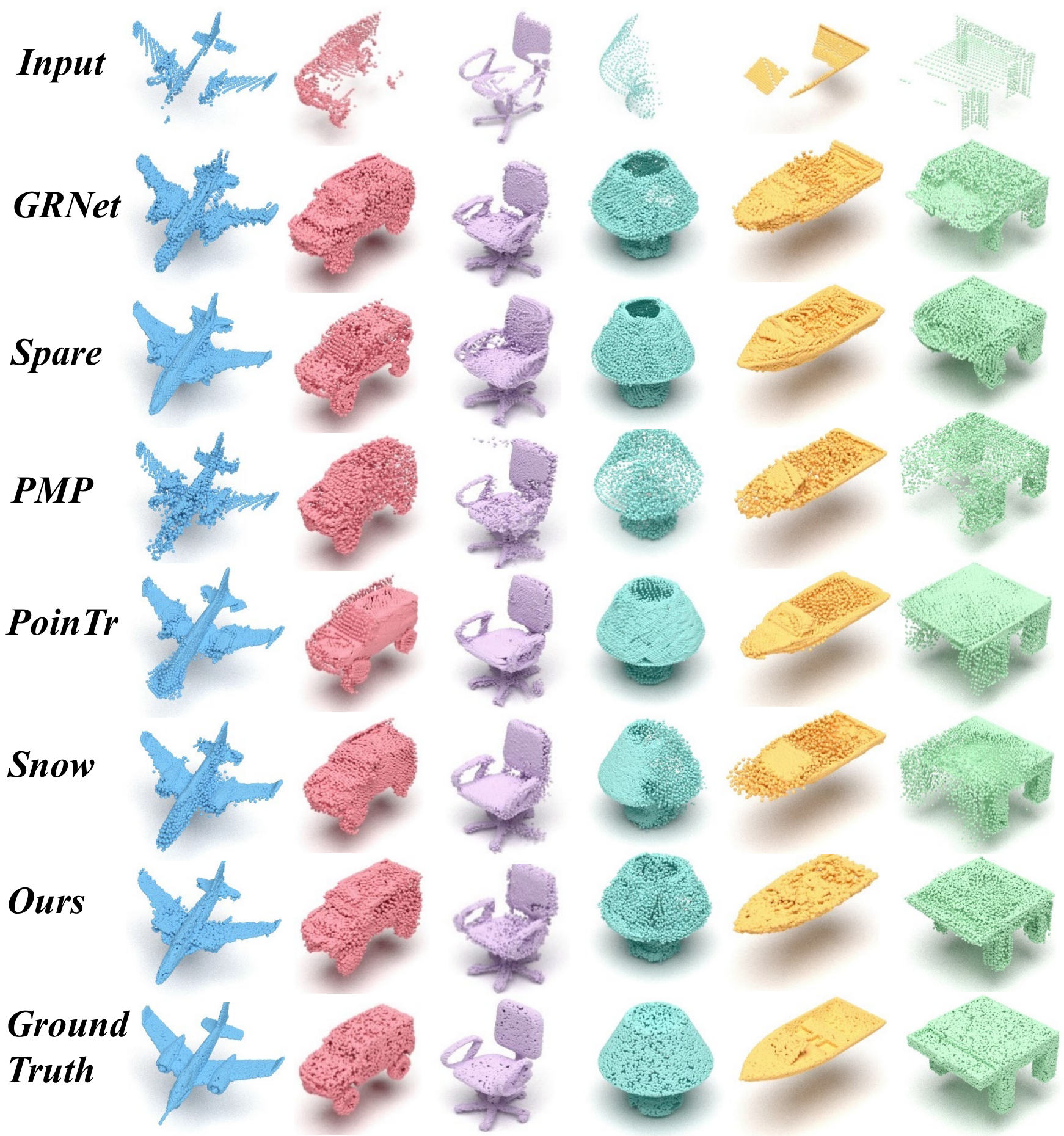}
    \caption{Visualization of point cloud completion comparison results on PCN dataset with other recent methods.}
    \label{fig:compare}
    \vspace{-4mm}
\end{figure}

\begin{table*}[htb!]
\centering\scriptsize
\setlength{\tabcolsep}{1.0mm}
\begin{tabular}{l|cc|cc|cc|cc|cc|cc|cc|cc|cc|cc|cc}
\toprule
Category & \multicolumn{2}{c|}{Bed} & \multicolumn{2}{c|}{Bench}  & \multicolumn{2}{c|}{Bookshelf} & \multicolumn{2}{c|}{FileCabinet} & \multicolumn{2}{c|}{Faucet} & \multicolumn{2}{c|}{Telephone} & \multicolumn{2}{c|}{Can} & \multicolumn{2}{c|}{Flowerpot} & \multicolumn{2}{c|}{Tower} & \multicolumn{2}{c|}{Pillow} & \multicolumn{2}{c}{Average}\\ \midrule
Metrics & CD & EMD & CD & EMD & CD & EMD & CD & EMD & CD & EMD & CD & EMD & CD & EMD & CD & EMD & CD & EMD & CD & EMD & CD & EMD\\ \midrule
Folding\protect~\cite{FoldingNet} & 3.17 & 73.6 & 1.45 & 50.1 & 2.48 & 64.4 & 1.94 & 65.3 & 3.19 & 66.2 & 0.69 & 39.1 & 1.76 & 60.2 & 4.11 & 82.9 & 1.83 & 59.5 & 1.64 & 63.2 & 2.06 & 60.2 \\
PCN\protect~\cite{PCN} & 2.50 & 49.4 & 0.96 & 28.9 & 2.39 & 44.7 & 1.49 & 37.2 & 1.96 & 40.3 & 0.54 & 24.0 & 1.30 & 30.9 & 2.58 & 48.7 & 1.34 & 33.7 & 1.09 & 31.5 & 1.36 & 34.0 \\
GRNet\protect~\cite{GRNet} & 0.93 & 29.7 & 0.86 & 25.8 & 0.93 & \textbf{29.6} & 1.57 & 24.9 & 0.83 & 27.6 &  0.87 & 26.2 & 1.15 & 32.3 & 1.24 & \textbf{33.5} & 0.87 & 25.3 & 1.06 & \textbf{28.8} & 1.15 & \textbf{28.2}\\
PoinTr\protect~\cite{Pointr} & 2.18 & 37.6 & 0.93 & 21.4 & 1.86 & 37.1 & 3.23 & 42.7 & 1.75 & 42.4 &  0.55 & \textbf{20.8} & 2.13 & 31.2 & 2.68 & 42.7 & 1.73 & 35.9 & 1.40 & 31.8 & 1.70 & 31.7\\\midrule
Ours & \textbf{0.72} & \textbf{28.4} & \textbf{0.71} & \textbf{18.2} & \textbf{0.89} & 29.7 & \textbf{0.97} & \textbf{16.4} & \textbf{0.34} & \textbf{20.5} & \textbf{0.48} & \textbf{20.8} & \textbf{0.63} & \textbf{29.0} & \textbf{1.19} & 35.9 & \textbf{0.60} & \textbf{21.9} & \textbf{0.97} & 29.8 & \textbf{0.89} & 31.0\\
\bottomrule
\end{tabular}
\vspace{-2mm}
\caption{Quantitative comparison results with other completion methods on ShapeNet55 dataset using CD-$l_2$($\times10^{3}$) and EMD($\times10^{3}$) metrics. We report the detailed results for each method on 10 sampled categories and overall average results on all 55 categories.}
\vspace{-5mm}
\label{tab:shape5}
\end{table*}

\begin{table}[]
\centering\scriptsize
\setlength{\tabcolsep}{0.6mm}
\begin{tabular}{l|cccccccc|c}
\toprule
CD-$l_2(\times10^{4})$ & Airplane & Cabinet & Car & Chair & Lamp & Sofa & Table & Vessel & Average \\ \midrule
Folding\protect~\cite{FoldingNet} & 3.151 & 7.943 & 4.676 & 9.225 & 9.234 & 8.895 & 6.691 & 7.325 & 7.142\\
PCN\protect~\cite{PCN} & 1.400 & 4.450 & 2.445 & 4.838 & 6.238 & 5.129 & 3.569 & 4.062 & 4.016\\
AtlasNet\protect~\cite{AtlasNet} &  1.753 & 5.101 & 3.237 & 5.226 & 6.342 & 5.990 & 4.359 & 4.177 & 4.523 \\
MSN\protect~\cite{MSN} & 1.543 & 7.249 & 4.711 & 4.539 & 6.479 & 5.894 & 3.797 & 3.853 & 4.758 \\
GRNet\protect~\cite{GRNet} & 1.531 & 3.620 & 2.752 & 2.945 & 2.649 & 3.613 & 2.552 & 2.122 & 2.723\\
PMP-Net\protect~\cite{PMP} & 1.205 & 4.189 & 2.878 & 3.495 & 2.178 & 4.267 & 2.921 & 1.894 & 2.878 \\
SpareNet\protect~\cite{Spare} & 1.756 & 6.635 & 3.614 & 6.163 & 6.313 & 7.893 & 4.987 & 3.835 & 5.149 \\
PointTr\protect~\cite{Pointr} & 0.993 & 4.809 & 2.529 & 3.683 & 3.077 & 6.535 & 3.103 & 2.029 & 3.345\\
Snowflake\protect~\cite{Snowflake} & 0.913 & 3.322 & 2.246 & 2.642 & \textbf{1.898} & 3.966 & 2.011 & 1.692 & 2.336\\\midrule
Ours & \textbf{0.646} & \textbf{2.594} & \textbf{1.743} & \textbf{2.149} & 2.759 & \textbf{2.186} & \textbf{1.876} & \textbf{1.602} & \textbf{1.944} \\ \bottomrule
\end{tabular}
\vspace{-3mm}
\caption{Quantitative comparison results with other methods of point cloud completion on PCN using CD-$l_2$ (lower is better). }
\vspace{-3mm}
\label{tab:pcncd}
\end{table}
\begin{table}[]
\centering\scriptsize
\setlength{\tabcolsep}{0.6mm}
\begin{tabular}{l|cccccccc|c}
\toprule
EMD($\times 10^{2}$) & Airplane & Cabinet & Car & Chair & Lamp & Sofa & Table & Vessel & Average \\ \midrule
Folding\protect~\cite{FoldingNet} & 1.682 & 2.576 & 2.183 & 2.847 & 3.062 & 3.003 & 2.500 & 2.357 & 2.526\\
PCN\protect~\cite{PCN} & 2.426 & 1.888 & 2.744 & 2.200 & 2.383 & 2.062 & \textbf{1.242} & 2.208 & 2.144\\
AtlasNet\protect~\cite{AtlasNet} & 1.324 & 2.582 & 2.085 & 2.442 & 2.718 & 2.829 & 2.160 & 2.114 & 2.282\\
MSN\protect~\cite{MSN} & 1.334 & 2.251 & 2.062 & 2.346 & 2.449 & 2.712 & 1.977 & 2.001 & 2.142 \\
GRNet\protect~\cite{GRNet} & 1.376 & 2.128 & 1.918 & 2.127 & 2.150 & 2.468 & 1.852 & 1.876 & 1.987\\ 
PMP-Net\protect~\cite{PMP} & 1.259 & 2.058 & 2.520 & 1.798 & \textbf{1.280} & 2.579 & 1.651 & 1.760 & 1.863\\ 
SpareNet\protect~\cite{Spare} & 1.131 & 2.014 & 1.783 & 2.050 & 2.063 & 2.333 & 1.729 & 1.790 & 1.862\\ 
PointTr\protect~\cite{Pointr} & \textbf{0.938} & 1.986 & 1.851 & 1.892 & 1.740 & 2.242 & 1.931 & 1.532 & 1.764\\
Snowflake\protect~\cite{Snowflake} & 1.375 & 2.633 & 2.591 & 2.086 & 1.599 & 3.070 & 1.616 & 1.957 & 2.116\\ \midrule
Ours & 0.958 & \textbf{1.830} & \textbf{1.564} & \textbf{1.667} & 1.782 & \textbf{1.755} & 1.499 & \textbf{1.402} & \textbf{1.557} \\ \bottomrule
\end{tabular}
\vspace{-2mm}
\caption{Quantitative comparison results with other methods of point cloud completion on PCN using EMD (lower is better).}
\vspace{-5mm}
\label{tab:pcnemd}
\end{table}

\subsection{Results on PCN dataset.}
We compare the performance of our proposed LAKe-Net and other state-of-the-art completion methods. We implement other methods using their open source code and hyper-parameters for fair comparison. Table~\ref{tab:pcncd} and~\ref{tab:pcnemd} show the quantitative comparison results of our method and other point cloud completion methods on PCN datasets, from which we can see that our method achieves the best performance over all counterparts on both CD and EMD metrics. Specifically, compared with the second-ranked Snowflake which also proposed progressive decoding modules, our method has better performance with the help of aligned keypoints and surface-skeletons. Besides, according to experimental results, our proposed LAKe-Net is more powerful to predict symmetrical geometries and their topology information compared with SnowflakeNet.

Moreover, we also show the visualization of qualitative comparison results and some recent methods in Figure~\ref{fig:compare}, which show that our method has better performance on completing missing topology. Specifically, methods which also utilize progressive coarse-to-fine decoding like PMP-Net and SnowflakeNet, tended to predict coarse missing shape and generate scattered points, especially for geometry with a plane or surface. Other methods like GRNet, SpareNet and PoinTr are weak on recovering the local details and some missing topology like table legs. Our method can predict geometries with more clear topology structure and fewer noises.

\subsection{Results on ShapeNet55 dataset}
\begin{figure}[]
    \centering
    \includegraphics[width=0.9\linewidth]{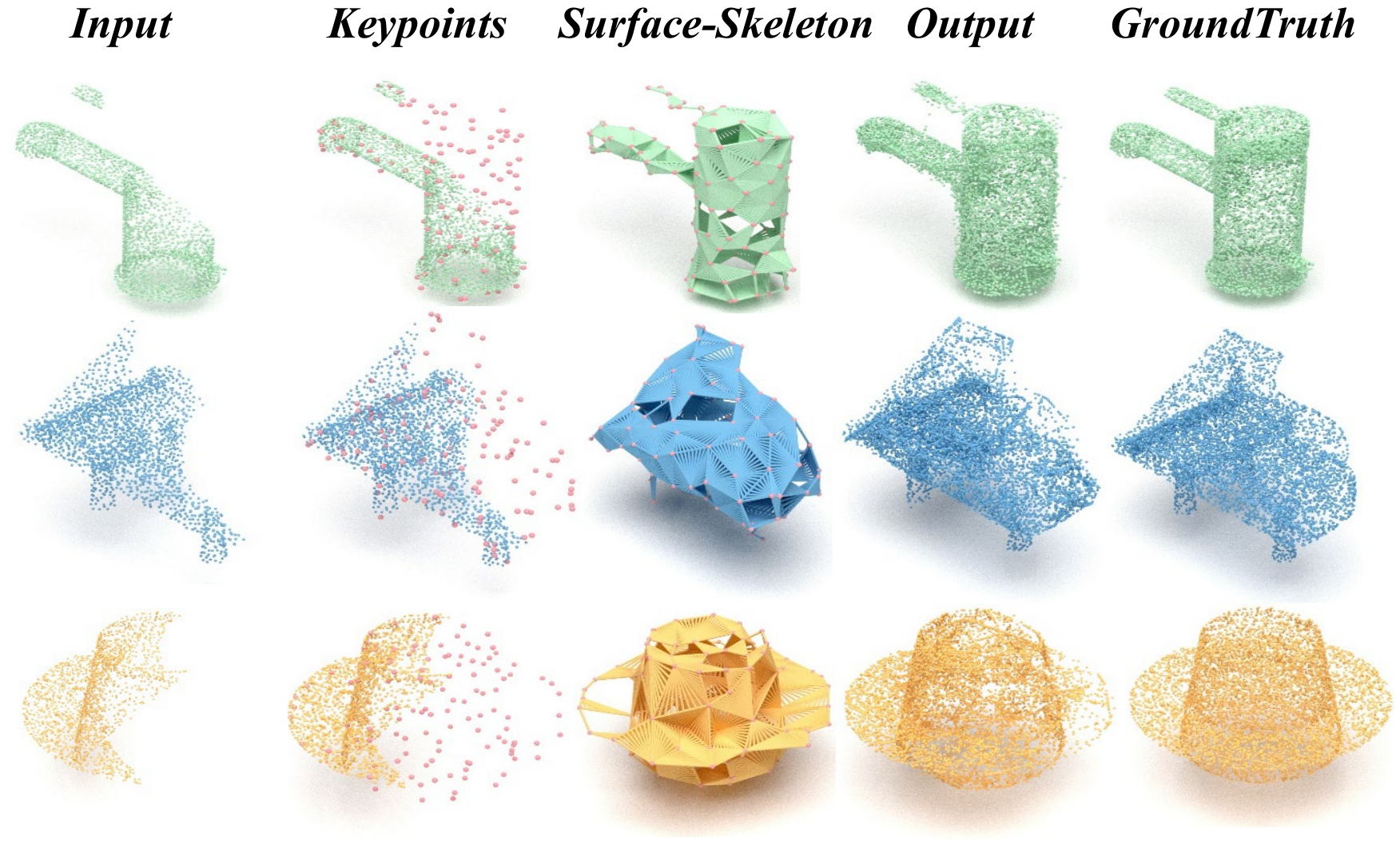}
    \vspace{-3mm}
    \caption{Visualization of completion on ShapeNet55 dataset by our proposed method. We also show the predicted keypoints and generated surface-skeletons on the second and third columns.}
    \vspace{-5mm}
    \label{fig:shapenet}
\end{figure}

Moreover, to evaluate the generalization and powerful of our method on a large account of categories of data to adapt to real-world scenarios, we conduct experiments on ShapeNet55 dataset and compare with other completion methods. We drop 75$\%$ of the complete point cloud and resample the remaining partial point clouds to 2,048 points as input for all methods.
Table~\ref{tab:shape5} shows the quantitative comparison results on 10 sampled categories. The last column shows the overall average results of 55 categories. We can see that our method achieves the best result on CD metric and have competitive results on EMD metric. Specifically, the results on \textit{Bed}, \textit{Bench}, \textit{Bookshelf}, \textit{FileCabinet}, \textit{Faucet}, which are similar as samples in PCN datasets or have shared topology within a sub-category, show that our method can recover geometries more efficiently using topology assistant. Moreover, the results on other categories show that our method is more powerful in completing geometries with regular and symmetrical contours, similar as \textit{Vessel} in PCN dataset. We visualize the completion process of our method on samples from ShapeNet55 in Figure~\ref{fig:shapenet}. It can be seen that our method can localize effect keypoints and recover missing topological and geometric information with the help of surface-skeletons.

\section{Method Analysis}
\vspace{-2mm}
In this section, we examine the effectiveness of our motivations in LAKe-Net. We conduct several ablation studies from different points of view. 
For fair comparison, all methods are trained and tested on PCN dataset for completion and KeypointNet for keypoint detection.

\textbf{Unsupervised Muti-scale Keypoint Detector.}
To prove the effectiveness and accuracy of our proposed UMKD, we evaluate our extracted keypoints compared with two recent unsupervised keypoint detectors Fernandez \textit{et al.}~\cite{fernandez2020unsupervised} and SkeletonMerger~\cite{SkeletonMerger}. All methods are trained and tested on KeypointNet~\cite{you2020keypointnet}, which has keypoints annotations with semantic correspondence labels. 
We evaluate these methods on five categories: airplane, car, chair, table and vessel. Specifically, we detect 16, 32, 64 keypoints for all categories. As for other methods, we train five models and detect 16 keypoints for each category. 
We firstly down-sample the same number of predict keypoints as the annotated keypoints using the nearest neighbor strategy.

\begin{figure}[]
    \centering
    \includegraphics[width=\linewidth]{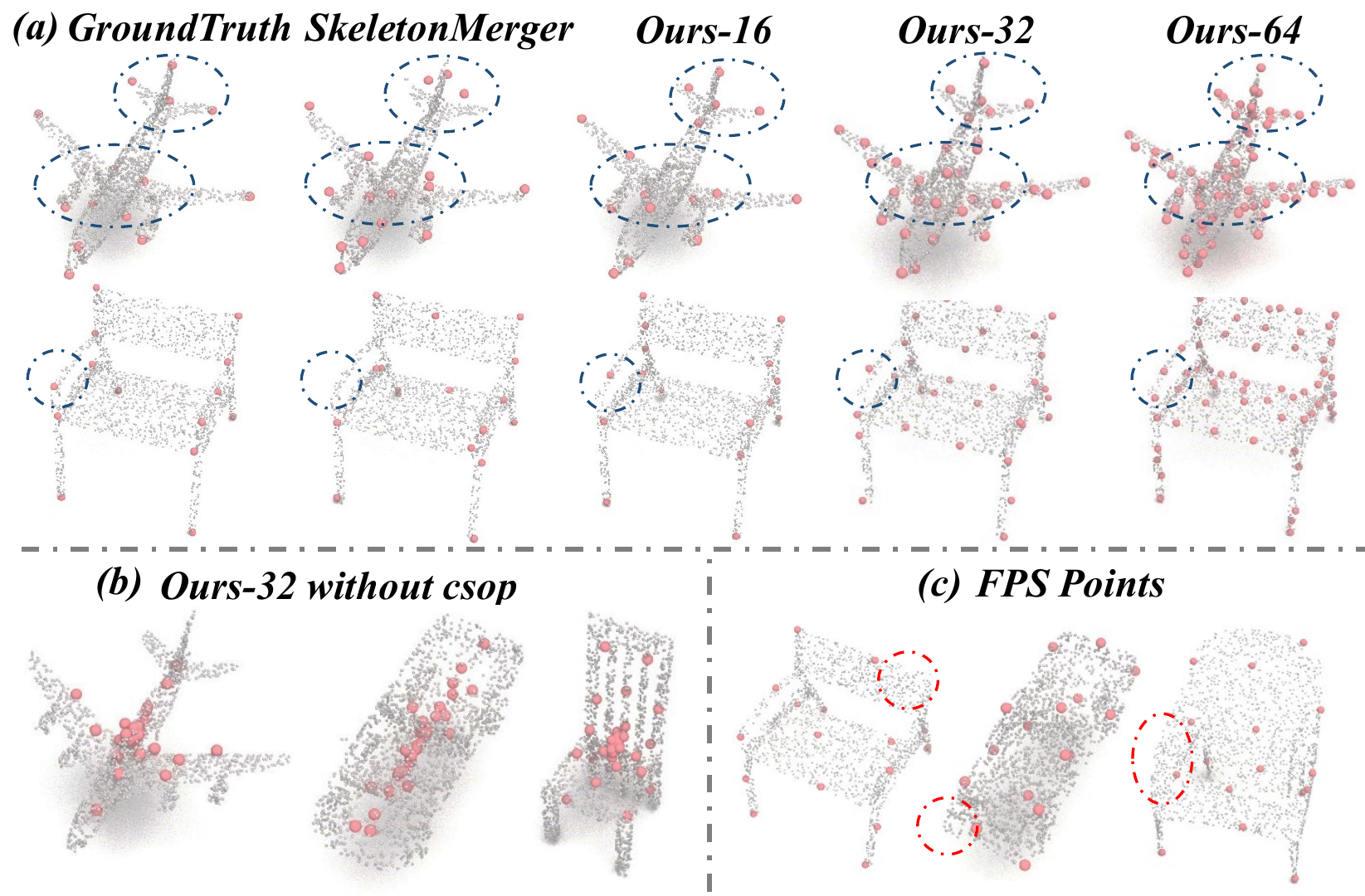}
    \vspace{-7mm}
    \caption{Visualization of (a) multi-scale keypoint detected by ours and SkeletonMerger on KeypointNet dataset; (b) failed case by our method without \textit{csop}; (c) drawbacks of FPS 16 points. \textit{csop} denotes category-specific offset predictors.}
    \vspace{-3mm}
    \label{fig:kp}
\end{figure}

For evaluation metrics, we follow SkeletonMerger and utilize mean Intersection over Unions (mIoU) metrics to evaluate the keypoint silence and accuracy.
It is calculated with a threshold of 0.1 using euclidean distance.
The quantitative results are shown in Table~\ref{tab:miou}, which illustrates that our proposed keypoint detector trained on multiple categories has competitive performance, or even better, than other unsupervised methods trained on a single category. The results on \textit{airplane}, \textit{chair} and \textit{table} also show that our method has better performance on geometries with obvious topological structures. The visualization is shown in Figure~\ref{fig:kp}(a). Our detector can produce more salient and semantic richer keypoints. We also show our detected keypoints in multi-scale which represent finer geometric details.

Besides, to evaluate the effectiveness of our proposed category-specific offset predictors (\textit{csop}) in UMKD, we replace all offset predictors with a single category invariant offset predictor. The qualitative results is visualized in Figure~\ref{fig:kp}(b) and quantitative results are shown in Table~\ref{tab:miou}. It is obvious that a single offset predictor cannot handle geometries in multiple categories. The predicted keypoints tend to aggregate together to reduce the loss.

\begin{table}[]
\centering\scriptsize
\setlength{\tabcolsep}{1.2mm}
\begin{tabular}{c|ccccc}
\toprule
mIoU & Airplane & Car & Chair & Table & Vessel\\\midrule
Fernandez \textit{et al.} & 69.7 & 50.5 & 51.2 & 49.3 & 53.5 \\
SkeletonMerger & 72.7 & 64.6 & 63.2 & 59.6 & 62.0 \\\midrule
Ours-16 & 73.2 & 58.1 & 69.2 & 62.5 & 61.3\\
Ours-32 & 73.7 & 60.2 & 70.5 & 63.2 & 62.5\\
Ours-64 & 74.0 & 62.9 & 71.3 & 65.4 & 64.0\\
Ours-32 \textit{w/o csop} & 35.4 & 13.2 & 27.8 & 23.9 & 11.0 \\
\bottomrule
\end{tabular}
\caption{Quantitative comparison results with other unsupervised keypoint detector on KeypointNet using mIoU (higher is better).}
\vspace{-7mm}
\label{tab:miou}
\end{table}

\textbf{Keypoints and Surface-skeletons.}
To evaluate the \rev{necessity of using aligned keypoints and surface-skeletons} for point cloud completion, we conduct several ablated experiments. We consider the auto-encoder $E_2$ and refinement subnet $R$ in the second stage as baseline.
In particular, we replace multi-scale complete keypoints detected in the first stage with multi-scale down-sampled points from ground truth using FPS strategy. And we use CD loss between down-sampled points and predicted keypoints in the second stage. We also remove the assist of generated surface-skeletons and change the up factors into [2,2,4] for fair comparison.
The quantitative results are illustrated in Table~\ref{tab:ablation}. We can see that the down-sampled points cannot represent efficient topology information (as shown in Figure~\ref{fig:kp}(c)) especially on some joint parts, and are not helpful for completing missing geometries in our pipeline. Besides, CD loss between two unordered and sparse point clouds is harder to be optimized to convergence. 

\begin{figure}
    \centering
    \includegraphics[width=0.9\linewidth]{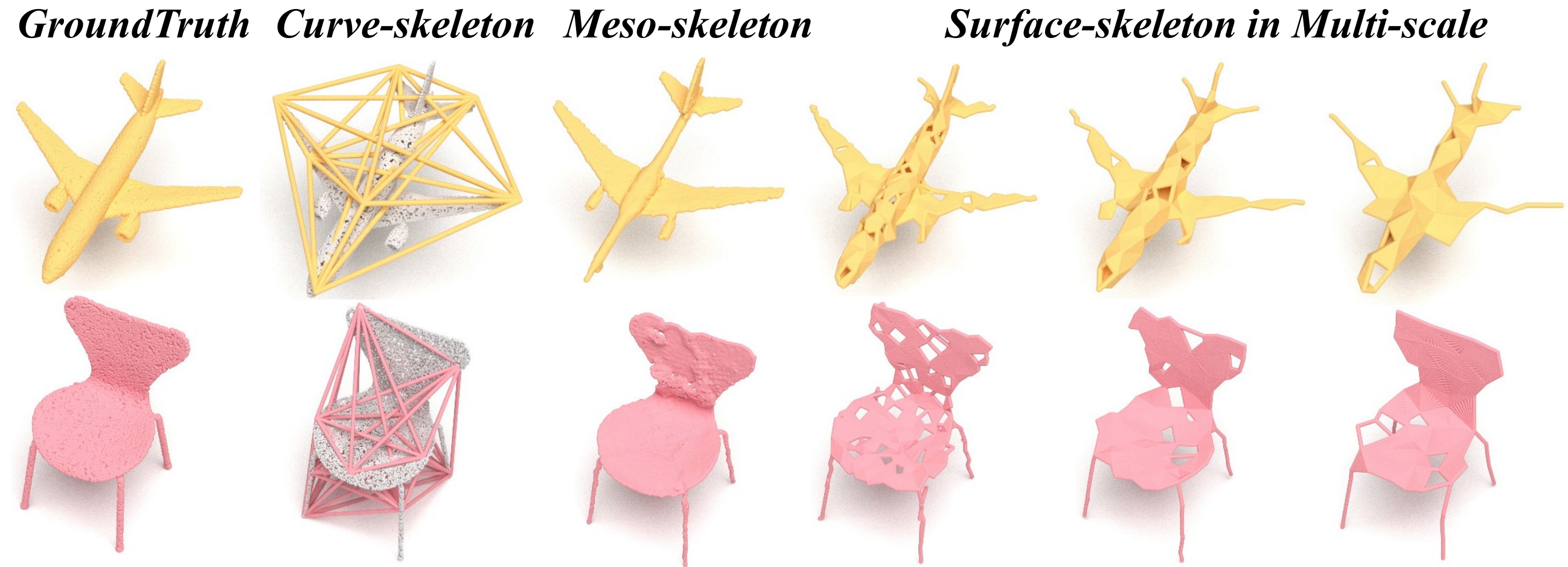}
    \vspace{-2mm}
    \caption{Visualization of different type of skeletons, including curve skeleton generated by~\cite{SkeletonMerger}, meso-skeleton from~\cite{DPC} and our surface-skeleton in multi-scale.}
    \vspace{-7mm}
    \label{fig:skeleton}
\end{figure}

We also visualize different types of skeletons compared with our multi-scale surface-skeleton in Figure~\ref{fig:skeleton}. It can be seen that our surface-skeleton focuses on representing surfaces of original shape, and can get competitive performance with meso-skeletons detected by typical method~\cite{DPC}, and better than curve skeleton from~\cite{SkeletonMerger}.

\begin{table}[]
\centering\scriptsize
\setlength{\tabcolsep}{0.6mm}
\begin{tabular}{l|cccccccc|c}
\toprule
EMD($\times 10^{2}$) & Airplane & Cabinet & Car & Chair & Lamp & Sofa & Table & Vessel & Average \\ \midrule

Ours & 0.958 & 1.830 & 1.564 & 1.667 & 1.782 & 1.755 & 1.499 & 1.402 & 1.557\\
~~-use FPS & 1.117 & 2.295 & 1.978 & 2.157 & 1.916 & 2.607 & 1.810 & 1.823 & 1.963 \\
~~-w/o S-sk & 1.469 & 2.638 & 2.386 & 2.380 & 2.221 & 2.989 & 1.906 & 2.020 & 2.251\\\midrule
PointDisturb & 1.031 & 1.902 & 1.554 & 2.012 & 1.945 & 2.037 & 1.684 & 1.437 & 1.700 \\
ClassDisturb & 0.963 & 1.846 & 1.576 & 1.786  & 1.831 & 1.780 & 1.545 & 1.397 & 1.590 \\
\bottomrule
\end{tabular}
\vspace{-3mm}
\caption{Ablation studies and robustness test on PCN dataset using EMD metrics. w/o S-sk denotes without surface-skeleton.}
\vspace{-2mm}
\label{tab:ablation}
\end{table}

\textbf{Robustness Test.}
We also conduct ablation studies to investigate the robustness of our method in some extreme cases. We firstly randomly disturb $5\%$ of the detected GT keypoints with a threshold of 0.1 after the first stage. Secondly, we deliberately misclassify geometries with simiar shapes: table, chair and sofa. The results are reported on Table~\ref{tab:ablation}. Experimental results show that our method is robust to errors in keypoints detection in the first stage.

\vspace{-3mm}
\section{Conclusion}
\vspace{-1mm}
In this paper, we propose a novel topology-aware point cloud completion method, named LAKe-Net, which focuses on completing missing topology by localizing aligned keypoints, with a novel Keypoints-Skeleton-Shape prediction manner, including aligned keypoints localization, surface-skeleton generation and shape refinement. 
Experimental results show that our LAKe-Net 
achieves the state-of-the-art performance on point cloud completion.

\vspace{-3mm}
\section{Acknowledgements}
\vspace{-1mm}
This work was supported by the National Key Research and Development Program of China (2019YFC1521104), National Natural Science Foundation of China (72192821, 61972157, 62176092), Shanghai Municipal Science and Technology Major Project  (2021SHZDZX0102), Shanghai Science and Technology Commission (21511101200, 22YF1420300, 21511100700), CAAI-Huawei MindSpore Open
Fund, and Art major project of National Social Science Fund (I8ZD22).

\newpage
{\small
\bibliographystyle{ieee}
\bibliography{egbib}
}

\end{document}